\pdfoutput=1

\documentclass[11pt]{article}

\usepackage{ACL2023}

\usepackage{times}
\usepackage{latexsym}
\usepackage{multirow}
\usepackage{numprint}
\usepackage[nameinlink]{cleveref}
\usepackage{graphicx}
\usepackage{svg}
\usepackage{arydshln}
\usepackage{xurl}

\usepackage[T1]{fontenc}
\usepackage[utf8]{inputenc}
\usepackage{booktabs}

\usepackage{microtype}

\usepackage{inconsolata}
\usepackage{tabularx}

\newcommand{\todo}[1]{{\color{red}}}

\title{Sheffield's Submission to the AmericasNLP Shared Task on Machine Translation into Indigenous Languages}

\author{ 
Edward Gow-Smith, Danae S\'{a}nchez Villegas
\\[0.3cm]
Computer Science Department, University of Sheffield, UK\\
{\small
{\tt \{egow-smith1, dsanchezvillegas1\}@sheffield.ac.uk}}\\
\\
}

\begin{document}
\maketitle
\begin{abstract}
In this paper we describe the University of Sheffield's submission to the AmericasNLP 2023 Shared Task on Machine Translation into Indigenous Languages which comprises the translation from Spanish to eleven indigenous languages. Our approach consists of extending, training, and ensembling different variations of NLLB-200. We use data provided by the organizers and data from various other sources such as constitutions, handbooks, news articles, and backtranslations generated from monolingual data. On the dev set, our best submission outperforms the baseline by 11\% average chrF across all languages, with substantial improvements particularly for Aymara, Guarani and Quechua. On the test set, we achieve the highest average chrF of all the submissions, we rank first in four of the eleven languages, and at least one of our submissions ranks in the top 3 for all languages.\footnote{We release code for training our models here: \url{https://github.com/edwardgowsmith/americasnlp-2023-sheffield}}

\end{abstract} 

\section{Introduction}
The 2023 AmericasNLP Shared Task \cite{ebrahimi-etal-2023-findings} involves developing machine translation systems for translating from Spanish to eleven low resource indigenous languages: Aymara (\textit{aym}), Bribri (\textit{bzd}), Ash\'{a}ninka (\textit{cni}), Chatino (\textit{czn}), Guarani (\textit{gn}), Wixarika (\textit{hch}), Nahuatl (\textit{nah}), Hñähñu (\textit{oto}), Quechua (\textit{quy}), Shipibo-Konibo (\textit{shp}), and Rar\'{a}muri (\textit{tar}). Developing machine translation systems for these languages is challenging since many of them are polysynthetic (i.e., words are composed of several morphemes) and word boundaries are not standardized; they present different orthographic variations (e.g., classical vs. modern Nahuatl variations); presence of code-switching is common, among other difficulties of low resource settings. 

Previous work has explored the effectiveness of pretrained machine translation models in low resource settings \cite{haddow-etal-2022-survey} showing their impact on improving translation quality and addressing data scarcity challenges. Following this approach, our submissions to the 2023 AmericasNLP shared task consist of extending and finetuning various versions of NLLB-200 \cite{costa2022no}, a state-of-the-art  machine translation model specifically designed for low resource settings. NLLB-200 is trained on \numprint{202} languages across \numprint{1220} language pairs, including three of the languages present in the AmericasNLP shared task: \textit{aym}, \textit{gn}, and \textit{quy}.\footnote{We present inference results on the dev set for these models in \Cref{tab:inference-dev}.} We further train our models on data from various sources such as constitutions and news articles, and we leverage multilingual training and ensembling to improve their performance. Models are evaluated using chrF  \cite{popovic-2015-chrf}, the official metric of the task. On the test set, we achieve the highest average chrF across all languages, and the best chrF for four of the languages.

The rest of the paper is organised as follows: \Cref{sec:data} describes the data sources for training our models, \Cref{sec:models} explains our three submissions in detail, \Cref{sec:results} presents the results on the dev and test sets, \Cref{sec:addexp} analyses the impact of different factors to the model's performance, \Cref{sec:zero-shot} looks at zero-shot capabilities, and we draw conclusions in \Cref{sec:conclusions}.

\begin{table*}[t!]
\centering
\small
\begin{tabular}{lrrrrrrrr}
\toprule
\textbf{Language} & \textbf{AmericasNLP 2023} & \textbf{Helsinki} & \textbf{REPUcs} & \textbf{NLLB} & \textbf{Train Total} & \textbf{Backtranslations} & \textbf{Bibles} \\
\midrule
aym & 15,586 & 149,225 & & 8,809 & 173,620 & 16,750 & 154,520 \\
bzd & 7,508 & & & & 7,508 &  & 38,502\\ 
cni & 3,883 & & & & 3,883 & 13,192 & 38,846 \\ 
czn & 3,118 & & & & 3,118 &  &  \\
gn & 26,032 & 1,713 & & 6,193 & 33,938 & 40,515 & 39,457 \\
hch & 8,966 & 2,404 & & & 11,370 & 510 & 39,756 \\
nah & 16,145 &3,848 & & & 19,993 & 8,703 & 39,772 \\
oto & 4,889 & 3,834 & & & 8,723 & 537 & 39,726 \\ 
quy & 542,914 & 3,634 &  10,729 & & 557,277 &  & 154,825 \\ 
shp & 14,592 & 14,656 & & & 29,248 & 23,592 & 79,341\\ 
tar & 14,721 & 3,856 & & & 18,577 &  & 39,444 \\
\bottomrule
\end{tabular}
\caption{Amount of parallel data collected for each language. AmericasNLP 2023: parallel training data provided by the organizers, Helsinski: data taken from \citet{vazquez-etal-2021-helsinki}, REPUcs: data taken from \citet{moreno-2021-repu}, NLLB: data from \citet{costa2022no}, Backtranslations: back-translations created from monolingual data, Bibles: data from The JHU Bible corpus \cite{mccarthy-etal-2020-johns}.}
\label{tab:data}
\end{table*}

\section{Data}
\label{sec:data}

\subsection{Data Collection}
We collect data from a variety of data sources, including training data provided by the organisers (AmericasNLP 2023), data from prior submissions to the AmericasNLP shared task (Helsinski and REPUcs) and relevant datasets specific to the indigenous languages included in the task (NLLB). \Cref{tab:data} shows the size of the training data for each language. The total amount of training data is unevenly distributed among datasets, with Quechua (\numprint{557277}), Aymara (\numprint{173620}), and Guarani (\numprint{33938}) having the greatest amount of training data. 

\paragraph{AmericasNLP 2023} Data provided by the organisers of the 2023 AmericasNLP Shared Task includes parallel datasets for training the eleven languages. \Cref{tab:anlp23data} contains all datasets and references.

\paragraph{Helsinski} We take data from OPUS \cite{tiedemann-2012-parallel} and other sources (including constitutions) provided by the University of Helsinski's submission \cite{vazquez-etal-2021-helsinki} to the AmericasNLP 2021 Shared Task \cite{mager-etal-2021-findings}. The collected data from constitutions includes translations of the Mexican constitution into Hñähñu, Nahuatl, Raramuri and Wixarika, of the Bolivian constitution into Aymara and Quechua, and of the Peruvian constitution into Quechua.

\paragraph{REPUcs} We use data collected for the REPUcs' submission to the 2021 AmericasNLP shared task \cite{moreno-2021-repu}. They introduce a new parallel corpus with Quechua data from three sources: (1) \citet{Duran}, which contains poems, stories, riddles, songs, phrases and a vocabulary for Quechua; (2) \citet{lyrics} which provides different lyrics of poems and songs; and (3) a Quechua handbook \cite{handbook}.

\paragraph{NLLB} We use two datasets introduced by \citet{costa2022no} as part of the training and evaluation for NLLB-200: (1) the NLLB Multi-Domain dataset, which provides \numprint{8809} English-Aymara examples in the news, health, and unscripted chat domains and (2) the NLLB Seed dataset, which contains \numprint{6193} English-Guarani examples consisting of professionally-translated sentences. 

\paragraph{Bibles} We also collect translations from the JHU Bible corpus \cite{mccarthy-etal-2020-johns}, which provides translations of the bible for all languages of the Shared Task except for Chatino. However, we do not observe performance improvements from using this data in our experiments (\Cref{sec:addexp}).

\begin{table*}[t!]
\centering
\small
\begin{tabular}{lcccccccccccc}
\toprule
\textbf{Model} & \textbf{aym} & \textbf{bzd} & \textbf{cni} & \textbf{czn} & \textbf{gn} & \textbf{hch} & \textbf{nah} & \textbf{oto} & \textbf{quy} & \textbf{shp} & \textbf{tar} & \textbf{mean} \\
\midrule
\textbf{Baseline}  & & &  &  &  &  &  &  &  & & & \\
\cite{vazquez-etal-2021-helsinki} & 32.7 & 23.8 & 26.8 & - & 31.1 & 29.9 & 29.8 & 14.7 & 33.8 & 31.7 & 19.6 & 27.4 \\ 
\midrule
\textbf{Submission 3} & & &  &  &  &  &  &  &  & & &   \\
NLLB-1.3B (single best) & 39.1 & 24.5 & 30.5 & 40.1 & 35.5 & 31.8 & 30.1 & 14.7 & 35.8 & 32.2 & 19.4 & 29.4 \\
\midrule
\textbf{Submission 2} & & &  &  &  &  &  &  &  & & &  \multirow{6}{*}{30.3} \\
NLLB-1.3B (best per lang) & & 24.6 & & & & & & & & & &  \\ 
NLLB-3.3B &  & & & & & & & & 38.8 & & &  \\ 
NLLB-1.3B (- NLLB Seed) & 41.1 & & & & & & & & & & & \\ 
NLLB-1.3B (+ backtrans 1) & & & & & 36.9 & & & & & & \\
NLLB-1.3B (+ backtrans 2) & & & & & & & & & & 35.4 & & \\ 
\midrule
\textbf{Submission 1} & & &  &  &  &  &  &  &  & & &  \multirow{6}{*}{30.5}  \\
Ensemble 1 & & 25.1 &  &  &  &  &  &  &  & & &   \\
Ensemble 2 & & & & 40.2 & & & & & & & & \\ 
Ensemble 3 & & & & & & 31.8 & & & & & & \\
Ensemble 4 & & & & & & & & & 39.1 & & & \\
Ensemble 5 & & & & & & & & & & & 20.0 & \\
\bottomrule
\end{tabular}
\caption{Dev set chrF scores for our three submissions. Here, the mean excludes \textit{czn}.}
\label{tab:dev-scores}
\end{table*}

\subsection{Backtranslations}
We generate backtranslations using the monolingual data sourced by \citet{vazquez-etal-2021-helsinki} for seven languages. This data comes from \citet{bustamante-etal-2020-data}, \citet{tiedemann2020tatoeba}, \citet{mager2018probabilistic}, \citet{tiedemann-2012-parallel}, and \citet{agic-vulic-2019-jw300}. We train NLLB-200 3.3B on X-\textit{es} for all \numprint{11} languages, X, in the task. We take two checkpoints of this model at different stages of training (\textbf{backtrans 1} and \textbf{backtrans 2}). We find this data improve performance for two of the languages in the task (\textit{gn} and \textit{shp}, see \Cref{sec:results}).

\subsection{Data Overlap}
We note that NLLB-200, the pretrained machine translation model we base our experiments on (see \Cref{sec:models}) is trained on a portion of the collected data. Specifically, Spanish-Aymara and English-Aymara data from GlobalVoices, and Spanish-Quechua data from Tatoeba, both as part of OPUS. We believe that the inclusion of this data will still be beneficial to the model, since NLLB-200 is not optimised for the languages we are interested in as part of this task. 

\subsection{Data Processing}
The training data provided by the organisers is tokenised for \textit{nah} and \textit{oto}. We detokenise it to put it in line with the rest of the training data. We replace punctuation not included in NLLB-200's vocabulary. For \textit{oto}, we find that 7\% of the dev set contains characters not in the vocabulary, since these characters do not occur in the training sets, we don't take steps to handle them. For \textit{czn}, we replace all superscript tone markings at the end of words with their standard counterparts, and then replace them naively back at inference.

\section{Models}
\label{sec:models}
To tackle the 2023 AmericasNLP task on automatic translation of eleven low resource indigenous languages, we use NLLB-200 \cite{costa2022no}, a state-of-the-art machine translation model specifically designed for low resource settings. We experiment with different distilled versions of NLLB-200 with 600M and 1.3B parameters, and the version with 3.3B parameters. Although inference results on three languages\footnote{NLLB-200 training data includes \textit{aym}, \textit{gn} and \textit{quy}.} show that the largest version, NLLB-3.3B, performs better than smaller versions (see \Cref{tab:inference-dev}), due to the large computational cost of using NLLB-3.3B we run most of our experiments with the 1.3B distilled version. Models are fine-tuned on all the training data (Train Total), i.e. all data sources in \Cref{sec:data} excluding Bibles and backtranslations, unless indicated. Moreover, we look at ensembling as an approach to improve the overall performance.

\begin{table*}[t!]
\centering
\small
\begin{tabular}{lcccccccccccc}
\toprule
\textbf{Submission} & \textbf{aym} & \textbf{bzd} & \textbf{cni} & \textbf{czn} & \textbf{gn} & \textbf{hch} & \textbf{nah} & \textbf{oto} & \textbf{quy} & \textbf{shp} & \textbf{tar} & \textbf{mean} \\
\midrule
3 & 35.3 & 24.5 & 28.5 & 39.9 & 39.1 & 32.0 & 27.3 & 14.8 & 37.2 & 28.6 & 18.4 & 29.6 \\
2 & 36.2 & 24.4 & &  & 39.3 & & & & 39.3 & 33.4 & & 30.3 \\
1 & & 25.0 & & 40.0 & & 32.3 & & & 39.5 &  & 18.7 & 30.5 \\ 
\bottomrule
\end{tabular}
\caption{Test set chrF scores for our three submissions. Here, the mean includes all languages.
}
\label{tab:test-scores}
\end{table*}

\begin{table}[t!]
\centering
\small
\begin{tabular}{lccc}
\toprule
\textbf{Model} & \textbf{quy} & \textbf{aym} & \textbf{gn} \\
\midrule
\textbf{Baseline} &&&\\
\cite{vazquez-etal-2021-helsinki}&33.8&32.7&31.1 \\
\midrule
\textbf{Inference} &&& \\ 
600M distilled & 30.0 & 34.2 & 32.5 \\
1.3B distilled & 31.0 & 35.2 & 35.2 \\ 
1.3B & 31.2 & 34.5 & 34.3 \\
3.3B & 32.9 & 35.4 & 35.6 \\ 
\bottomrule
\end{tabular}
\caption{\label{tab:inference-dev} Dev set chrF results for various NLLB-200 models, compared to the baseline and our submissions.}
\end{table}

\begin{table}[t!]
\centering
\small
\begin{tabular}{lc}
\toprule
\textbf{Trained} & \textbf{quy} \\
\midrule
Everything & 37.8 \\
Decoder & 36.5 \\
3 Decoder Layers & 34.2 \\
\bottomrule
\end{tabular}
\caption{\label{tab:inference-dev} Dev set chrF results for various NLLB-200 models, compared to the baseline and our submissions.}
\end{table}
37.8
34.2
35.2

\paragraph{Submission 3} We train NLLB-200 1.3B distilled on the training data\footnote{We exclude Bibles data and backtranslations.} and we choose the best checkpoint based on average chrF across all languages. We submit translations for all languages using this model (\textbf{NLLB-1.3B (single best)}).

\paragraph{Submission 2} We take the best-performing single model per language, excluding ensembles. We find that for the majority of languages, the best single model (by dev chrF) is the same as Submission 3, so we only submit additional translations for five languages: 
\begin{itemize}
    \item \textbf{NLLB-1.3B (- NLLB Seed) - \textit{aym}} NLLB-1.3B trained on all data (Train Total) except for NLLB Seed.
    \item \textbf{NLLB-1.3B (best per lang) - \textit{bzd}} NLLB-1.3B trained on all data.
    \item \textbf{NLLB-1.3B (+ backtrans 1) - \textit{gn}} NLLB-1.3B trained on all data plus backtranslations from checkpoint 1.
    \item \textbf{NLLB-3.3B - \textit{quy}} NLLB-3.3B trained on all data.
    \item \textbf{NLLB-1.3B (+ backtrans 2) - \textit{shp}} NLLB-1.3B trained on all data plus backtranslations from checkpoint 2.
\end{itemize}

\paragraph{Submission 1} We experiment with various ensembles of models in attempt to improve performance further -- we only find improvements over Submission 2 through ensembling for five of the languages in the task. These selected ensembles are as follows:
\begin{itemize}
    \item \textbf{Ensemble 1 - \textit{bzd}} The best NLLB-1.3B model for \textit{bzd} and an NLLB-600M model trained on all languages.
    \item \textbf{Ensemble 2 - \textit{czn}} The best average NLLB-1.3B model and an NLLB-3.3B model trained on all languages.
    \item \textbf{Ensemble 3 - \textit{hch}} The best average NLLB-1.3B model and an NLLB-600M model trained on all languages.
    \item \textbf{Ensemble 4 - \textit{quy}} NLLB-3.3B trained on all languages, NLLB-3.3B trained on just the three supported languages (\textit{aym}, \textit{gn}, and \textit{quy}), and NLLB-1.3B trained on all languages.
    \item \textbf{Ensemble 5 - \textit{tar}} NLLB-1.3B trained on all languages, NLLB-600M trained on all languages, and NLLB-1.3B trained on all languages with a label smoothing of 0.2 (rather than 0.1).
\end{itemize}

\subsection{Experimental Setup}
We train the models in a multilingual fashion across all \numprint{11} language pairs present in the task, extending the embedding matrix to cover the tags for the new languages. We experiment with freezing various parameters, but find best results from training everything. We run our experiments on a single A100 GPU with batch sizes of 64, 16, and 2 for the 600M-, 1.3B-, and 3.3B-parameter models, respectively. We run our experiments in fairseq \cite{ott2019fairseq}. Full hyperparameters for all of our runs are provided in \Cref{tab:hyperparameters}. To evaluate our models, following the official evaluation, we use chrF \cite{popovic-2015-chrf} computed using SacreBLEU \cite{post-2018-call} with signature: \texttt{\tiny nrefs:1|case:mixed|eff:yes|nc:6|nw:0|space:no|version:2.1.0}.

\begin{table*}[hbt!]
\centering
\small
\begin{tabular}{lcccccccccccc}
\toprule
\textbf{Model} & \textbf{aym} & \textbf{bzd} & \textbf{cni} & \textbf{czn} & \textbf{gn} & \textbf{hch} & \textbf{nah} & \textbf{oto} & \textbf{quy} & \textbf{shp} & \textbf{tar} & \textbf{mean} \\
\midrule
NLLB-1.3B (single best) & 39.1 & 24.5 & 30.5 & 40.1 & 35.5 & 31.8 & 30.1 & 14.7 & 35.8 & 32.2 & 19.4 & 30.3 \\
NLLB-3.3B only quy & & & & & & & & & 35.3 & & & \\
NLLB-3.3B all langs & & & & & & & & & 38.3 & & & \\ 
1.3B random initialisation & 21.9 & 17.6 & 24.2 & 33.7 & 22.8 & 25.1 & 24.3 & 13.7 & 22.9 & 22.2 & 16.9 & 22.3 \\
NLLB-1.3B + bibles & 38.3 & 24.1 & 30.0 & 38.0 & 35.5 & 30.0 & 28.0 & 14.7 & 35.2 & 31.9 & 18.9 & 28.7 \\
\bottomrule
\end{tabular}
\caption{\label{table:extra-experiments} Dev set chrF scores for our additional experiments. For comparison, we reproduce the best single model as the first row.}
\end{table*}

\section{Results}
\label{sec:results}

\subsection{Dev Set Results}

\Cref{tab:dev-scores} presents the results of our models on the dev set. We observe that for all languages, at least one of our models outperforms the baseline \cite{vazquez-etal-2021-helsinki}, with the exception of \textit{oto} where we obtain comparable performance. The greatest improvements over the baseline model are on the three NLLB supported languages: \textit{aym} (41.1 compared to 32.7), \textit{gn} (36.9 compared to 31.1) and \textit{quy} (39.1 compared to 33.8). We note that backtranslations only lead to improved performance on \textit{gn} and \textit{shp}, which are the two languages with the greatest amount of available monolingual data.

\paragraph{Inference results} NLLB-200 is trained on data from three of the languages in this shared task: \textit{quy}, \textit{aym}, \textit{gn}. \Cref{tab:inference-dev} shows the inference results for these languages on the dev set for different variations of NLLB-200 models, along with our submissions. We observe a considerable improvement from the distilled 600M to 1.3B distilled models, with the greatest improvement over the baseline model for \textit{gn}. We note that the 1.3B and 3.3B models outperform the baseline model for \textit{aym} and \textit{gn}. For \textit{quy}, the inference results are worse than the baseline, likely due to the large amount of training data available in the task. We are able to improve substantially upon the inference results for \textit{quy} and \textit{aym}, but much less so for \textit{gn} -- this may be due to much less training data being available for \textit{gn} compared to the other two languages. 

\subsection{Test Set Results}
Results on the test set are shown in \Cref{tab:test-scores}. Overall, our best submission achieves the highest average chrF across all languages from all submissions to the task (the second-best average is 29.4, compared to our 30.5). We also rank first for four of the eleven languages: \textit{aym}, \textit{czn}, \textit{quy}, and \textit{shp}. Our biggest improvement upon the second-place team is for \textit{czn}, where we achieve 40.0 compared to 36.6. Submissions 1 and 2 rank in the top 3 for all languages. Surprisingly, the best chrF score was obtained on \textit{czn} (40.0), the language with the least amount of training data (\numprint{3118} examples), followed by \textit{quy} (39.5), and \textit{aym} (36.5).

\section{Additional Experiments}
\label{sec:addexp}
We provide the results of additional experiments to better understand the impact of various factors to our model's performance. The results of these experiments are shown in \Cref{table:extra-experiments}. 

\paragraph{Multilingual training} We look into whether multilingual training is beneficial to the model. For this, we train a 3.3B-parameter model on the \textit{quy} data only, and compare this version (NLLB-3.3B only quy) to the one trained on all languages (NLLB-3.3B all langs) at the same number of updates (\numprint{480000}). We find that multilingual training greatly improves the performance on \textit{quy}, suggesting the model benefits from transfer learning across the languages. We suspect the benefit of the multilingual approach is related to the fact that although the languages included in the task are from different linguistic families, they share linguistic properties (e.g., polysynthetic or agglutinative).

\paragraph{Random initialization} To analyse the benefit of starting from NLLB-200, we train an equivalent model to the 1.3B parameter version with randomly-initialised parameters. We see that this model performs much worse than the equivalent NLLB-200 model. As expected, we observe the greatest differences on the languages supported by NLLB-200 (\textit{aym}, \textit{gn}, \textit{quy}).

\paragraph{Bibles data} Similar to findings of \citet{vazquez-etal-2021-helsinki}, we observe a drop in average performance through training on the Bibles data for the majority of languages except for \textit{gn} and \textit{oto}, where we obtain comparable performance.

\section{Zero-shot Performance}
\label{sec:zero-shot}

We investigate whether our models have any zero-shot capabilities, i.e. translating a language pair for which the model has not seen any training data. For this, we take the best-performing model for \textit{es}-\textit{shp} (NLLB-1.3B + backtrans 2), and evaluate it on translating \textit{quy}-\textit{shp}, \textit{aym}-\textit{shp}, and \textit{gn}-\textit{shp}.\footnote{This is possible due to multiparallel dev sets across all languages.} The results of these experiments are shown in \Cref{tab:zero-shot}. We find that our model is able to retain decent performance for these three zero-shot directions (maximum 25\% drop in chrF), despite training all of the parameters of the machine translation model.

\begin{table}[t!]
\centering
\tiny
\begin{tabular}{lcccc}
 & \textbf{es-shp} & \textbf{quy-shp} & \textbf{aym-shp} & \textbf{gn-shp} \\
\midrule
NLLB-1.3B (+ backtrans 2) & 35.4 & 30.5 & 29.3 & 26.7 \\
\bottomrule
\end{tabular}

\caption{\label{tab:zero-shot} Dev set chrF scores for three zero-shot translation directions with our best model for \textit{es-shp}.}
\end{table}

\section{Conclusions}
\label{sec:conclusions}
In this paper we describe our submissions to the AmericasNLP 2023 Shared Task. We participated with three submissions which consist of training different versions of the NLLB-200 model on publicly available data from different sources. Models are trained in a multilingual fashion and we experiment with different ensembles of models to further improve performance. We improve upon the inference scores for NLLB-200 3.3B for its three supported languages, and our best submission achieved the highest average chrF across all languages of any submission to the task.

\section*{Acknowledgments}
This work is supported by the Centre for Doctoral Training in Speech and Language Technologies (SLT) and their Applications funded by the UK Research and Innovation grant EP/S023062/1.

\bibliography{anthology,custom}
\bibliographystyle{acl_natbib}

\clearpage
\appendix

\section{Hyperparameters}\label{sec:app-hypeparameters}
\begin{table}[h!]
    \small
    \centering
    \begin{tabular}{cc}\toprule
         \textbf{Hyper-parameter} & \textbf{Value}  \\
         \midrule
         Batch size & 16$^\dagger$ \\
         Update freq & 1 \\
         Max learning rate & 0.01 \\
         Schedule & inverse square root \\
         Warmup steps & \numprint{10000} \\
         Adam betas & 0.9, 0.98 \\
         Label smoothing & 0.1$^\ddagger$ \\
         Weight decay & 0.0001 \\
         Dropout & 0.3 \\
         Clip norm & 1e-6 \\
         Language pair temperature & 3$^\star$ \\
         Number of updates & 1M \\
         Valid freq & 40K updates + every epoch \\
         Beam size & 5 \\\bottomrule
    \end{tabular}
    \caption{Hyper-parameters used to train our models. \\$\dagger$: 64 for NLLB-600M, 2 for NLLB-3.3B. \\$\ddagger$: 0.2 for one of our models, used in Ensemble 5. \\$\star$: 1 for  NLLB-3.3B models (including for backtranslations)}
    \label{tab:hyperparameters}
\end{table}

\begin{figure}[hbt!]
    \includegraphics[width=\columnwidth]{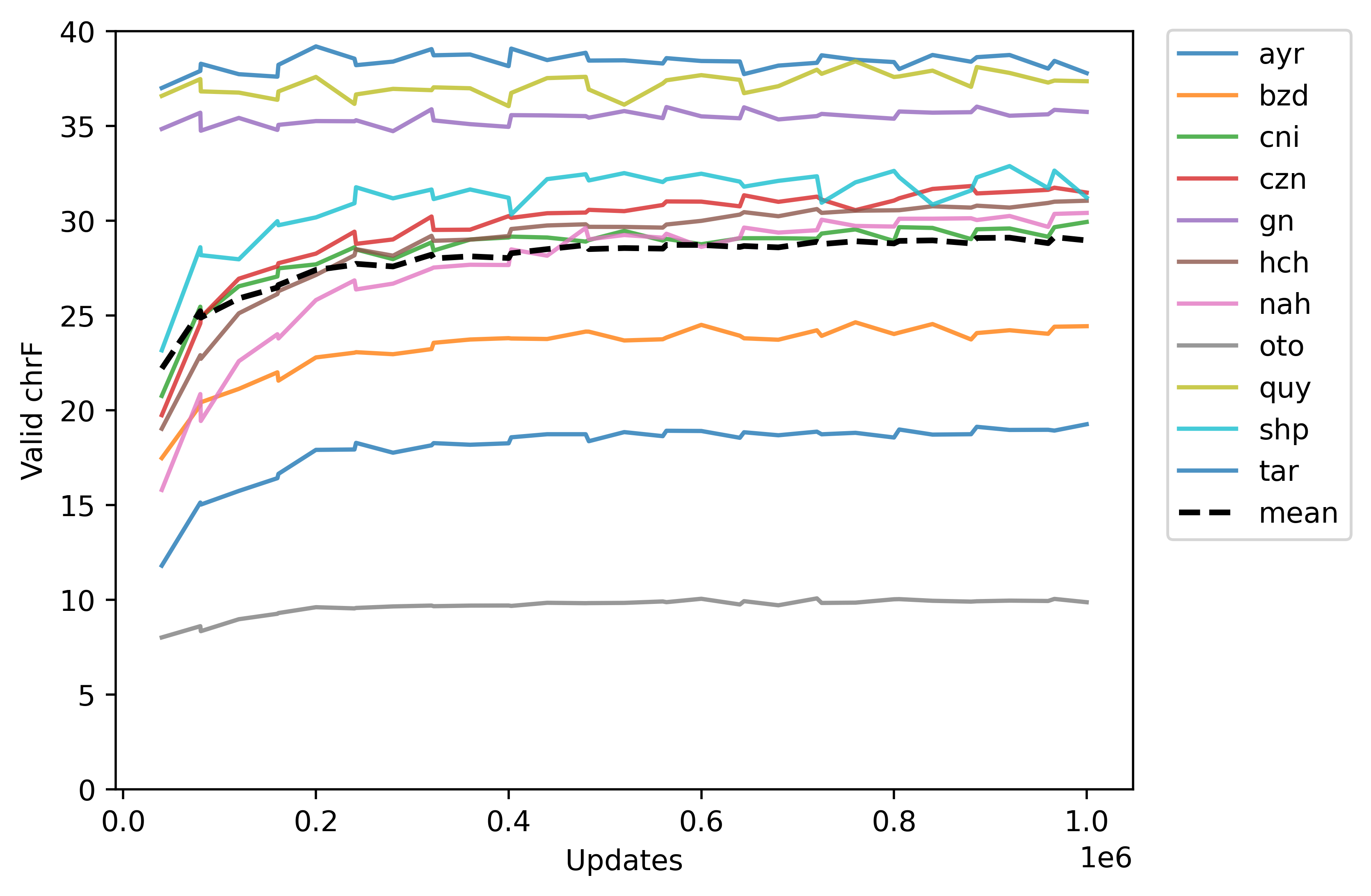}
    \caption{Valid chrF scores during training of our best single model (Submission 3).}
    \label{fig:chrf_curves}
\end{figure}

\begin{figure}[hbt!]
    \centering
    \includegraphics[width=\columnwidth]{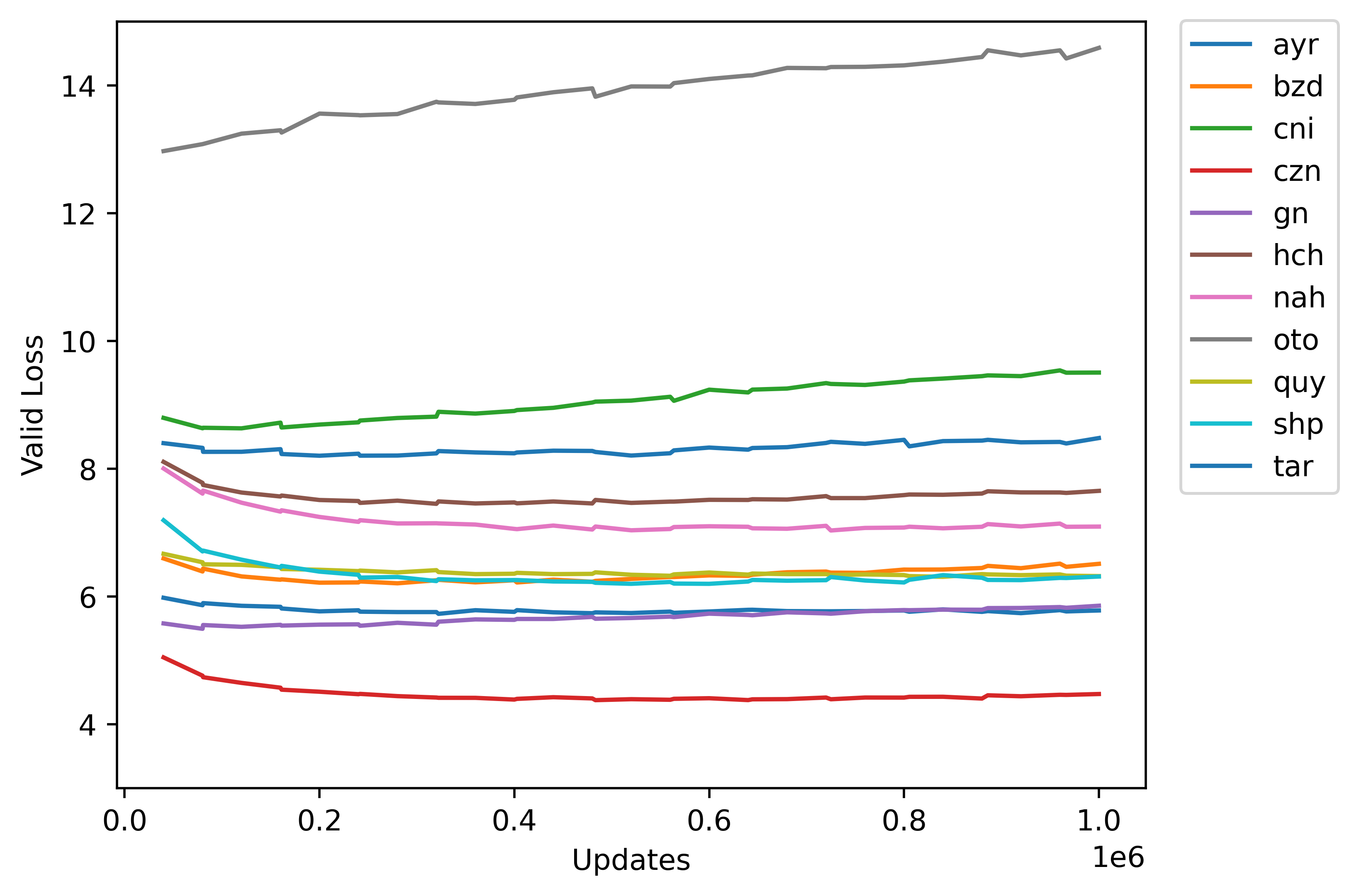}
    \caption{Valid losses during training of our best single model (Submission 3).}
    \label{fig:loss_curves}
\end{figure}

\begin{table}[t!]
\centering
\tiny
\begin{tabular}{cp{4.8cm}}
\toprule
\textbf{Dataset}                         & \textbf{Source}                                                                     \\ 
\midrule
 & \citet{ortega-etal-2020-overcoming}                                                 \\ 
 & \citet{cushimariano:prel:08}                                                        \\ 
\multirow{-3}{*}{ashaninka-spanish}      & \citet{mihas:anaani:11}                                            \\ 
\midrule
aymara-spanish                           & GlobalVoices \cite{tiedemann-2012-parallel}                                                     \\ 
\midrule                                         & \citet{adolfo1998curso}                                                             \\ 
                                         & \citet{solorzano2017corpus}                                                         \\ 
                                         & \citet{jara2018gramatica}                                                           \\ 
                                         & \citet{murillo2013se}                                                               \\ 
                                         & \citet{jara1993tte}                                                                 \\ 
\multirow{-6}{*}{bribri-spanish}         & \citet{enrique2005diccionario}                                                      \\
\midrule
guarani-spanish                          & \citet{chiruzzo-etal-2020-development}                                              \\ 
\midrule
hñähñu-spanish                           & Tsunkua \url{https://tsunkua.elotl.mx/about/}                                               \\ 
\midrule
wixarika-spanish       & \citet{mager2020wixarika}                                                           \\
\midrule
                                         & \citet{montoya-etal-2019-continuous}                                               \\ 
\multirow{-3}{*}{shipibo\_konibo-spanish} & \citet{galarreta-etal-2017-corpus}                                            \\ 
\midrule
raramuri-spanish                         & \citet{brambila1976diccionario}                                                     \\
\midrule
                          & JW300 \cite{agic-vulic-2019-jw300}                                                       \\ 
\multirow{-2}{*}{quechua-spanish}                                       & GlobalVoices \cite{tiedemann-2012-parallel}                                                     \\
\midrule
nahuatl-spanish                          & Axolotl \cite{gutierrez-vasques-etal-2016-axolotl}                                         \\ 
\midrule
chatino-spanish          & IUScholar Works  \\
 & \url{https://scholarworks.iu.edu/dspace/handle/2022/21028} \\
\hline
\end{tabular}
\caption{Data provided by the organisers of the 2023 AmericasNLP Shared Task.}
\label{tab:anlp23data}
\end{table}

\end{document}